\documentclass[smallextended]{svjour3}       
\smartqed  
\usepackage{lmodern}
\usepackage{graphicx}
\usepackage{gb4e}
\usepackage{url}
\usepackage{hyperref}
\usepackage{amsmath}
\usepackage[T1]{fontenc}
\usepackage{natbib}
\usepackage{longtable}
\usepackage{enumerate}

\usepackage{latexsym}

\newcommand*{\affmark}[1][*]{\textsuperscript{#1}}

\begin{document}


\title{An Annotated Commodity News Corpus for Event Extraction}


%



\author{Meisin Lee\protect\affmark[1]  \and
        Lay-Ki Soon  \and
        Eu-Gene Siew  \and 
        Ly Fie Sugianto }


\institute{All authors are from Monash University \\
              \email{\{mei.lee, soon.layki, siew.eu-gene, lyfie-sugianto\}@monash.edu} \\      
}

\date{Received: date / Accepted: date} 

\maketitle

\begin{abstract}
Commodity news contain a wealth of information; one of the key information is the analysis of recent commodity price movements along with notable events that led to that movement. Through event extraction, useful information can be extracted from commodity news. Besides the details of price movements, causal relation between events and commodity price movement can be extracted as well. To facilitate future research in this area, we present a new dataset with the following information identified and annotated: (i) entities (both nominal and named), (ii) events (trigger words and argument roles), (iii) event metadata: modality, polarity and intensity,  and (iv) event-event relations. These information, when extracted, can be very beneficial for understanding the correlation of current world affairs and commodity price movement, which can then be used for commodity price prediction. To the best of our knowledge, this is the first corpus that is annotated with elements which are crucial for event extraction from commodity news.

\end{abstract}


\section{Introduction} \label{intro}
Financial markets are sensitive to breaking news on economic events. Specifically for crude oil markets, it is observed in \citep{brandt2019macro} that news about macroeconomic fundamentals and geopolitical events affect the price of the commodity. Therefore accurate and timely automatic identification of events in news items is crucial for making timely trading decisions. 
Commodity news contain a wealth of information. Generally, news of this genre contain summaries or recent updates of the commodity in focus. The analyses by the reporter or journalist of past events provide a good distillation of world events that are truly causal to the movement of commodity prices. Understanding the correlation between events and commodity price movement can be useful for other downstream tasks such as for the purpose of learning event sequence (also known as scripts in \citep{schank2013scripts}) and for predicting commodity price movement. Commodity news typically contain these few key information: (i) analysis of recent commodity price movements (up, down or flat), (ii) a retrospective view of notable causal event(s) that led to the movement, (iii) forecast or forward-looking analysis of supply-demand situation as well as projected commodity price targets, and (iv) events metadata. 

Below are five sentences taken from a number of commodity news articles, illustrating the various types of information found in the commodity corpus:

\begin{exe}
		\ex U.S. crude stockpiles \textbf{soared}(E1) by 1.350 million barrels in December from a mere 200 million barrels to 438.9 million barrels, due to this \textbf{oversupply}(E2) crude oil prices \textbf{plunged}(E3) more than 50\% as Tuesday.\footnote{this example will be used throughout to the paper to illustrate the annotation details.}
		\ex Brent crude oil \textbf{climbed}(E4) towards \textdollar 112 a barrel on Thursday on fears that a slower global economy will demand \textbf{less}(E5) fuels than anticipated. 
		\ex Oil is cruising \textbf{higher}(E6) after OPEC+ refused to \textbf{cut}(E7) supplies. 
		\ex Crude oil futures \textbf{fell}(E8) in Asian trading on Wednesday after the International Energy Agency \textbf{trimmed}(E9) its demand forecast for the fourth quarter of this year.
		\ex It is expected that London-traded crude future to \textbf{drop}(E10) to the \textdollar 78 level by the beginning of the second quarter since OPEC is considering a reduction in its supply  \textbf{cuts}(E11).
	\end{exe}

The types of information can be categorized into the following groups of key information:
\begin{enumerate}[i]
    \item \textbf{Analysis of recent commodity price movement} - Events E1, E3, E4, E6, E8, and E10 are recent commodity price movements.
    \item \textbf{Retrospective view of notable causal event(s) that led to commodity price movement} - Events E2, E5, E7, E9, and E11 are causal events that led to the the corresponding price movement. Details of causal event relation is covered in Section \ref{subsection:eventRelation}.
    \item \textbf{Forecast or forward-looking analysis of supply-demand situation or price targets} - Event E9 is an example of forecast of demand, while event E10 is an example of anticipated commodity price target.
    \item \textbf{Event Metadata}:
        \begin{itemize}
            \item \textbf{Modality}: Event E5 is an anticipated event that has yet to take place.
            \item \textbf{Polarity}: Event E7 did not take place since it is being \textit{refused}.
            \item \textbf{Intensity}: Event E11 is an event that has taken place but intensity is reduced. \\
            Detailed description of Event Metadata is covered in Section \ref{subsection:metadata}
        \end{itemize}
\end{enumerate}

Specifically for this genre of text, it is insufficient to conduct Event Detection without extracting event arguments as well. To illustrate this point, consider event E3 in example sentence (1), trigger word "plunged", it is incomplete to just identify the event \textbf{movement-down-loss} without knowing "what" \textit{plunged}. In commodity news, possible events are \textit{`price plunged'}, \textit{`supply plunged'} and \textit{`demand plunged'}. As seen sentence (1), the sentence consists of two movement events (E1 and E3). To disambiguate and to extract events accurately, it is important to extract event arguments accurately as well.

After an extensive search of conference papers and journals, we have come to a conclusion that there is no publicly available annotated commodity news corpus that is suitable for event extraction. Our contribution to facilitate future research in this area is a new annotated commodity\footnote{At this stage, we target only commodity news relating to \textbf{crude oil}} dataset, characterising 18 event types (Appendix \ref{appendix:eventTypes}), with 21 entity types (Appendix \ref{appendix:entityTypes}) along with each events' metadata: \textit{polarity}, \textit{modality} and \textit{intensity}. Each event type has its own set of argument roles, in total there are 20 argument roles (Appendix \ref{appendixEventArguments}). This paper describes the annotation methodology used in the commodity news corpus and makes the annotated corpus available for academic research purpose\footnote{\href{https://github.com/meisin/Commodity-News-Corpus}{https://github.com/meisin/Commodity-News-Corpus}}. 

In this paper, information about the corpus, including data collection and annotation procedure are laid out in Section \ref{sec:Corpus}. Annotation methodologies for \textbf{Entities} and \textbf{Events} (Event Trigger, Event Arguments, Event Metadata and Event Relation) are covered in detail in Section \ref{sec:Annotation}. Annotation evaluation and challenges are stated in Section \ref{sec:challenges}. Potential uses of the corpus are covered in Section \ref{sec:Uses}; it is then followed by Conclusion in Section \ref{sec:Conclusion}.

\section{Related Work}
The annotation methodologies presented here are conceived based on the notions proposed by canonical programs such as ACE2005 (Automatic Content Extraction)
, TAC-KBP
Event track shared task and ERE (Entities, Relations and Events). An extensive comparison has been made in \citep{aguilar2014comparison}, where authors analyzed and provided a summary of the different annotation approaches. Subsequently there were a number of works that expanded earlier annotation standards, such as in \citep{ogorman-etal-2016-richer}, authors introduced the Richer Event Description (RED) corpus and methodologies that annotate entities, events, times, entities relations (coreference and partial coreference), and events relations (temporal, causal, and sub-events). On the other hand, authors in \citep{thompson2017enriching} introduced a more comprehensive way of annotating event metadata, which is known as \textit{meta-knowledge} in the paper, covering \textit{modality}, \textit{subjectivity}, \textit{source}, \textit{polarity} and \textit{specificity} of the event. Their schema was tailored for news events. In \citep{araki-etal-2018-interoperable}, the authors 
presents methodologies for annotating events and event relations across different domains, as an extension to formalization focusing on domain-specific event ontology as seen in ACE2005 and TAC-KBP.

\subsection{Financial Events} \label{subsec:financeResources}
In the domain of Finance and Economics, it is found that annotated datasets for the task of event extraction are rather scarce. Most of the available datasets are on company-related events and are used mainly for extracting company events for company stock price prediction. Examples of such datasets are: 

\begin{itemize}
    \item Dataset released in \citep{malik2011accurate} is a dataset focused on financial events found in European corporate press releases for events such as \textit{dividends}, \textit{profits announcements}, and other quantitative events (non-quantitative events such as mergers and acquisition are excluded). The paper proposed the use of statistical classifiers aided by rules to extract these events. 
    \item Dataset in \citep{hogenboom2013semantics} is a dataset consisting of ten different company-specific financial events; they are \textit{announcements regarding CEOs, presidents, products, competitors, partners, subsidiaries, share values, revenues, profits,} and \textit{losses}. Along with the dataset, authors presented an event detection framework basing on knowledge-bases and patterns and relying on domain ontology. The authors used rule-sets or domain ontology knowledge-bases that are mostly or entirely created by hand. 
    \item Dataset in \citep{jacobs2018economic} is a dataset consisting of ten types of company-economic events which were manually annotated: \textit{Buy ratings, Debt, Dividend, Merger \& acquisition, Profit, Quarterly results, Sales volume, Share repurchase, Target price, Turnover}. Event detection was done via the supervised classification approach on this dataset.
    \item In \citep{dor2019financial}, the authors proposed using Wikipedia sections to extract weak labels for sentences describing company events to overcome the constraint of using predefined event taxonomies.
    \item Authors in \citep{yang-etal-2018-dcfee} presented a system that can automatically generate a large scale labeled data and extract financial events at document-level from Chinese Financial News. The experiments were carried out on four types of financial events: \textit{Equity Freeze, Equity Pledge, Equity Repurchase,} and \textit{Equity Overweight}. 
\end{itemize}
All existing resources listed above are focused on \textbf{company events} and they cannot be directly used in our work because company-related events are essentially different from macro-economic and geo-political events found in commodity news. After an extensive search for commodity news related resources, we came across RavenPack's\footnote{RavenPack is an analytics provider for financial services. Among their services are finance and economic news sentiment analysis. More information can be found on their page: https://www.ravenpack.com/} crude oil dataset. This dataset is available through subscription at the Wharton Research Data Services (WRDS). It is made up of news headlines and a corresponding sentiment score generated by Ravenpack's own analytic engine. Unfortunately this dataset is not suitable for the task of event extraction as it only contains sentiment score without any annotation on events. However, Ravenpack's event taxonomy on crude oil-related events proves to be a useful resource in helping us define our own event taxonomy. Details of event taxonomy is covered in Section \ref{subsection:eventTaxanomy}.

\section{Corpus Description} \label{sec:Corpus}
In this section, we describe the corpus and the data collection process in detail.

\subsection{Data Collection}
The commodity news articles were collected from the following news agencies\footnote{Due to copyright issues, original articles are not released. Instead URL links to each articles are provided along with their corresponding annotation file.}:
\begin{itemize}
    \item \href{https://www.investing.com/commodities/crude-oil-news}{https://www.investing.com/commodities/crude-oil-news}
    \item \href{https://www.reuters.com/news/economy}{https://www.reuters.com/news/economy}
    \item \href{https://www.cnbc.com}{https://www.cnbc.com}
    \item \href{https://www.hellenicshippingnews.com/}{https://www.hellenicshippingnews.com/}
    \item \href{https://oilprice.com/Latest-Energy-News/}{https://oilprice.com/Latest-Energy-News/}
    \item \href{https://www.marketwatch.com}{https://www.marketwatch.com}
    \item \href{https://www.marketpulse.com}{https://www.marketpulse.com}
    \item \href{https://www.fxempire.com/news/}{https://www.fxempire.com/news/}
\end{itemize}

From the pool of commodity news articles, about 150 pieces of news articles were chosen based on their headlines. This ensures we create a balanced dataset with each event type represented as evenly as possible to avoid imbalanced data. More information about event types is included in Section \ref{subsection:eventTaxanomy}.

\subsection{Annotation Procedure} \label{subsec:procedure}
\begin{figure}[h]
\centering
  \includegraphics[width=\textwidth]{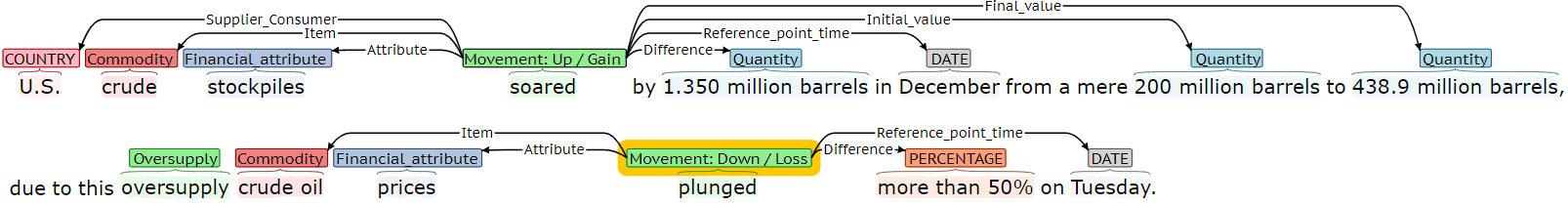}
\caption{Annotation using Brat annotation tool: (i) entities (both nominal and named) are annotated with entity types listed above the respective words (in various colours except green), (ii) events trigger words are also annotated (in green), and (iii) entities are linked to their respective event trigger through arches, argument roles these entities play in linked events are listed on the arches. \\ Note: Event metadata: modality, polarity and intensity, and  event-event relations are not shown in this diagram.}
\label{fig:annotation}       
\end{figure}

The dataset is annotated using \textbf{Brat rapid annotation tool} \citep{stenetorp2012brat}, a web-based tool for text annotation. The annotated version of Sentence (1) using Brat is shown in Figure \ref{fig:annotation}. 

The annotation process is designed to have high inter-annotator agreement (IAA). One of the criteria is that annotators should possess basic knowledge in business, finance, and economics. As such, annotators were chosen from a pool of students from Monash School of Business. Annotators were then given annotation training and provided with clear annotation schemas and examples. Every piece of text was duly annotated by two annotators independently. 

The annotation was done based on the following layers:
\begin{itemize}
    \item Layer 1: Identify and annotate entity mentions.
    \item Layer 2: Annotate events by identifying event triggers.
    \item Layer 3: Using event triggers as anchors, identify and link surrounding entity mentions to their respective events. Annotate the argument roles each entity mention plays with respect to the events identified. 
    \item Layer 4: Annotate event metadata: modality, polarity and intensity. Each Modality, Polarity and Intensity label will be supported by the existence a corresponding cue word. 
    \item Layer 5: Annotate event-event relations.  
\end{itemize}

After each layer, an adjudicator assessed the annotation and evaluated inter-annotator agreement before finalizing the annotation. For cases where there are annotation discrepancies, the adjudicator will act as the tie-breaker to decide on the final annotation. Once finalized, annotators then proceed with the next layer. This is done to ensure no accumulation of the previous layer's errors in the subsequent layers of annotation.

\section{Annotation Methodologies} \label{sec:Annotation}
As described in Section \ref{subsec:procedure}, annotation is done layer by layer at the sentence level.
This section describes our definition of events and principles for annotation of entity mentions and events. Annotation of events is further divided into (i) annotating entity mentions, (ii) annotating event triggers, (iii) linking entity mentions to their respective events and identifying the argument roles each entity mention plays, (iv) assigning the right metadata labels to \textit{Polarity}, \textit{Modality} and \textit{Intensity}, and (iv) annotating event-event relations. 

\subsection{Entity Mention}
An entity mention is a reference to an object or a set of objects in the world, including named entities, nominal entities, and pronouns. For simplicity and convenience, \textbf{values} and \textbf{temporal expressions} are also considered as entity mentions in this work. There are 21 entity types identified and annotated in the dataset. Nominal entities relating to Finance and Economics are annotated. Apart from crude oil-related terms, below here are some examples of nominal entities found in the Commodity News Corpus and was duly annotated:
\begin{enumerate}
    \item attributes : \textit{price, futures, contract, imports, exports, consumption, inventory, supply, production, usage, deliveries}
    \item economic entity : \textit{economic growth, economy, market(s), economic outlook, growth, dollar, green back, employment data}
\end{enumerate}
The list of entity types and the corresponding list of entities (both nominal and named entities) are listed in Appendix , the list of entity types are listed in Appendix \ref{appendix:entityTypes}Se\ref{appendix:entityTypes}.


\subsection{Events}
Events are defined as `specific occurrences', involving `specific participants'\footnote{\href{https://www.ldc.upenn.edu/sites/www.ldc.upenn.edu/files/english-events-guidelines-v5.4.3.pdf}{https://www.ldc.upenn.edu/sites/www.ldc.upenn.edu/files/english-events-guidelines-v5.4.3.pdf}}. The occurrence of an event is marked by the presence of an event trigger. In addition to identifying triggers, all of the participants of each event are also identified. An event's participants are Entities that are involved in that Event. Details and rules for identifying event triggers are described in Section \ref{subsection:eventTriggers}, while details for identifying Event Arguments are covered in Section \ref{subsection:eventArguments}.

\subsubsection{\textbf{Event Triggers}} \label{subsection:eventTriggers}
Our annotation of Event Triggers is aligned to TAC-KBP Event Guidelines and also to \citep{mitamura2015event} where an event trigger (known as event nugget in the shared task) can be either a single word  (main verb, noun, adjective, adverb) or a  continuous or discontinuous multi-word phrase. Here are some examples found in the dataset: 
\begin{itemize}
    \item \textbf{Verb}: Houti rebels \textbf{attacked} Saudi Arabia.
    \item \textbf{Noun}: The government slapped \textbf{sanctions} against its petroleum....
    \item \textbf{Adjective}: A fast \textbf{growing} economy has...
    \item \textbf{Multi-verb}: The market \textbf{bounced back}....
\end{itemize}
We have defined a set of 18 oil-related event types as our annotation scope. Event types and the corresponding list of example trigger words are listed out in Table \ref{table:EventTypes} in Appendix \ref{appendix:eventTypes}. 

\subsubsection{Event Taxonomy} \label{subsection:eventTaxanomy}
According to \citep{brandt2019macro} who analyzed the Ravenpack's sentiment score of each event type and oil price, events that move commodity prices are \textbf{geo-political}, \textbf{macro-economic} and \textbf{commodity supply and demand} in nature. The list of Ravenpack's Event Taxonomy as analyzed by \citep{brandt2019macro} is found in Appendix \ref{appendix:eventTaxanomy}. Based on Ravenpack's event taxonomy, 18 types of events coupled with event arguments can be further grouped into the following categories:
\begin{itemize}
    \item \textbf{geo-political:} Geo-political tension, civil unrest, embargo / sanctions, trade tensions and other forms of geo-political crisis (event details and example sentence in Appendix \ref{appendix:eventTypes}.1).
    \item \textbf{macro-economic:} US employment data, economic / gross domestic product (GDP) growth, economic outlook, growth forecast (event details and example sentence in Appendix \ref{appendix:eventTypes}.2).
    \item \textbf{commodity supplies and demand:} Demand / supply decrease, demand / supply increase, action taken to increase demand / supply and action taken to increase demand / supply (event details and example sentence in Appendix \ref{appendix:eventTypes}.3 and \ref{appendix:eventTypes}.4).
    \item \textbf{commodity price movement:} price increase, price decrease, price forecast (event details and example sentence in Appendix \ref{appendix:eventTypes}.5). 
\end{itemize}

\subsubsection{Event Arguments} \label{subsection:eventArguments}
\begin{figure}[h]
\centering
  \includegraphics[width=\textwidth]{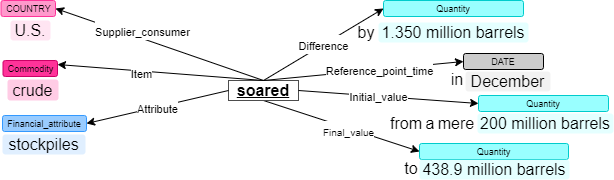}
\caption{The first event from Figure \ref{fig:annotation} is re-drawn here to show the details clearly. Event trigger (\textbf{soared}) is linked to its event arguments with argument roles displayed on the arches.}
\label{fig:partAnnotation}       
\vspace{-0.7em}
\end{figure}

After event triggers and entity mentions are annotated, they need to be linked up to form events. An event contains an event trigger and a set of event arguments. Details of events and their corresponding list of arguments are found in Appendix \ref{appendixEventArguments}. Referring to Figure~\ref{fig:partAnnotation}, the event trigger \textbf{soared} is linked to seven entity mentions via arches. The argument role of each entity mention is labeled on each arch respectively, while entity types are labeled in various colours on top of each entity span. This information is also summarized in Table~\ref{table:EventArguments}. 

 \begin{table}[h!]   
    \centering
    \caption{List of Event Arguments of example shown in Figure \ref{fig:partAnnotation}}.
    \begin{tabular}{ p{2.5cm} | p{2cm}  | p{6cm}}  \hline
    \textbf{Argument Role} & \textbf{Entity Type} & \textbf{Sentence} \\ \hline
    Supplier & Country & \textbf{[U.S.]} crude.....\textbf{soared} by... \\ \hline
    Item & Commodity & U.S. \textbf{[crude]}.....\textbf{soared} by... \\ \hline
    Attribute & Financial-Attribute & U.S. crude \textbf{[stockpiles]} \textbf{soared} by...\\ \hline
    Difference & Quantity & ....\textbf{soared} by \textbf{[1.350 million barrels]}...\\ \hline
    Reference time & Date & ....\textbf{soared} by ....in \textbf{December}... \\ \hline
    Initial value & Quantity & ...\textbf{soared} by...from a mere \textbf{200 million barrels}... \\ \hline
    Final value & Quantity & ...\textbf{soared} by...to \textbf{438.9 million barrels}. \\ \hline
    \end{tabular}
    \label{table:EventArguments}
\end{table}

\subsubsection{Event Metadata: Event Polarity, Modality and Intensity} \label{subsection:metadata}
After events are identified, they are also assigned a label each for Event Polarity, Modality, and Intensity. Cue words for Polarity, Modality, and Intensity are identified and annotated. The presence of these cue words acts as supporting evidence when assigning event metadata labels. Each metadata is described in detail in the following subsections. 

\paragraph{POLARITY} (Possible labels for Event Polarity are: POSITIVE and NEGATIVE) \\ An event has the value POSITIVE unless there is an explicit indication that the event did not take place, in which case NEGATIVE is assigned.

\begin{exe}
    \item OPEC countries \textit{refused} to \textbf{cut}_{caused-movement-down-loss} oil supplies. \\
    Polarity: NEGATIVE (cue: refused)
    \item  OPEC's efforts \textit{prevented} oil prices from \textbf{rising}_{movement-up-gain} much.
    Polarity: NEGATIVE (cue: prevented) 
\end{exe}


\paragraph{MODALITY} (Possible labels for Event Modality are: ASSERTED and OTHER) \\ Event modality determines whether the event represents a ``real'' occurrence. ASSERTED is assigned if the author or speaker refers  to it as though it were a real occurrence, and OTHER otherwise. OTHER covers believed events, hypothetical events, commanded and requested event, threats, proposed events, discussed events, desired events, promised events, and other unclear construct.

\begin{exe}
    \item The market \textit{expects} US to \textbf{sanction}_{embargo} Iran. \\
    Modality: OTHER (cue: expects) 
    \item Analysts were \textit{anticipating} oil inventories to \textbf{fall}_{movement-down-loss} by 800,000. \\
    Polarity: OTHER (cue: anticipating) 
\end{exe}

\paragraph{INTENSITY} (Possible labels for Event Intensity are: NEUTRAL, INTENSIFIED and EASED) \\
Event intensity is a new event metadata, specifically created for this work to better represent events found in this corpus. Oftentimes, events described in Commodity News are about the intensity of an existing event, whether the event is further intensified or eased. 

Examples of events that are INTENSIFIED:
\begin{exe}
    \item ...could hit Iraq 's output and \textit{deepen} a supply \textbf{shortfall}_{undersupply}.\\
    Intensity: INTENSIFIED (cue: deepen)
\end{exe}

Examples of events that EASED:

\begin{exe}
    \item Libya 's civil \textbf{strife}_{civil-unrest} has been \textit{eased} by potential peace talks.\\
    Intensity: EASED (cue: eased)
\end{exe}
The event \textbf{strife} (civil unrest) in sentence (11) is not an event with negative polarity because the event has actually taken place but with reduced intensity. INTENSITY label is used to capture the interpretation accurately, showing that the civil unrest event has indeed taken place but with less intensity. 

With these three event metadata information, we can annotate and capture all essential information about an event. To further illustrate this point, consider the list of examples of complex events below:

\begin{exe}
    \item \textit{Delay} a \textit{planned} \textit{easing} of output \textbf{cuts}$_{caused-movement-down-loss}$ \\
    Polarity: NEGATIVE (cue: delay) \\
    Modality: OTHER (cue: planned) \\
    Intensity: EASED (cue: easing) 
    \item In order to end the global crisis, OPEC may \textit{hesitate} to implement a \textit{planned} \textit{loosening} of output \textbf{curbs}_{caused-movement-down-loss}. \\
    Polarity: NEGATIVE (cue: hesitate)\\
    Modality: OTHER (cue: planned)\\
    Intensity: EASED (cue: loosening) 
    \item Oil prices rose to \$110 a barrel on rumours of a \textit{renewed} \textbf{strife}_{civil-unrest}. \\
    Polarity: POSITIVE  \\
    Modality: OTHER (cue: rumours) \\
    Intensity: INTENSIFIED (cue: renewed)
\end{exe}

\subsubsection{Event Relation} \label{subsection:eventRelation}
In \citep{araki-etal-2018-interoperable} and in \citep{o2016richer}, various event-event relations are identified and annotated. For this work, however, only the \textbf{causal-preconditions} relation is annotated. This is because, among the relationships, the one that is of interest is the cause-effect relationship of events, i.e. the events that caused commodity price movement. Similar to \citep{araki-etal-2018-interoperable} and \citep{o2016richer}, two types of causal-preconditions identified here are:

\vspace{-0.7em}
\paragraph{Explicit causal-preconditions}: There are two types of explicit causality: (i) one with explicit connective: with the verbs such as “because” which have obvious causal meanings, and (ii) one with ambiguous connectives with causal meaning such as “due to”. This is illustrated in sentence (15) with causal connectives underlined.

\begin{exe}
    \item Prices have \textbf{fallen}_{movement-down-loss} from \$115 \underline{due to} an oil \textbf{glut}_{oversupply}, prices \textbf{rebounded}_{movement-up-gain} \underline{aided by} sharp \textbf{fall}_{movement-down-loss} in U.S. oil drilling.
\end{exe}

\paragraph{Implicit causal-preconditions}: 
Implicit causality is often expressed in absence of causal connectives in the text, or there is only one entity of the cause or effect. For simplicity, in this work, both prior events and simultaneous events are considered as implicit causal-precondition, as shown in sentence (16) and (17): 

\begin{exe}
    \item Oil price \textbf{dropped}_{movement-down-loss} \underline{as} supply \textbf{increased}_{movement-down-loss}.
    \item Oil prices pushed to new \textbf{high}_{position-high} \underline{amid} supply \textbf{worries}_{negative-sentiment}.
\end{exe}

\subsection{Data Augmentation}
To increase the size of the annotated dataset, we introduced more variance through data augmentation: (i) trigger word replacement and (ii) event argument replacement).

\subsubsection{Trigger word replacement}
FrameNet was utilized to augment available data and to generate both diverse and valid examples. FrameNet\footnote{\href{https://framenet.icsi.berkeley.edu}{https://framenet.icsi.berkeley.edu}} shares with ACE/ERE a goal of capturing information about events and relations in text. Authors in \citep{aguilar2014comparison} pointed out that all events, relations, and attributes that were represented by ACE2005 or ERE and TAC-KBP standards can be mapped to FrameNet representations through some adjustments, as shown in Table~\ref{table:mappingACE}:
\vspace{-0.7em}
\begin{table}[h!]   
    \centering
    \caption{The mapping between ACE schemas and FrameNet frames}
    \begin{tabular}{ p{3.5cm} | p{3.5cm} }  \hline
    \textbf{ACE Schema} & \textbf{FrameNet Frame} \\ \hline
    Event Trigger & Lexical Unit \\ \hline
    Event Arguments & Frame Elements \\ \hline
    Argument Role & Name of Frame Elements \\ \hline
    Event Type & Frame \\ \hline
    Entity & Entity \\ \hline
    \end{tabular}
    \label{table:mappingACE}
    \vspace{-0.7em}
\end{table}

A number of annotated sentences were selected, and their trigger words were replaced with words (known as \textbf{lexical units} in FrameNet) of the same frame in FrameNet. The idea is to replace the existing trigger word with another valid trigger word while maintaining the same semantic meaning (in FrameNet's term - maintaining the same frame). Event trigger replacement is illustrated in sentence (18):
\begin{exe}
    \item The benchmark for oil prices \textbf{advanced} 1.29\% to \$74.71. \\
    Potential trigger word replacements are \textbf{surged}, \textbf{rose}, \textbf{appreciated}, \textbf{climbs}, etc.
\end{exe}

\subsubsection{Event argument replacement}
Apart from data augmentation through trigger word replacement, at the same time, we have also replaced event arguments. The replacement candidates were chosen from a pool of candidates of the same entity type and the same argument role within the pool of existing annotations, as illustrated in sentence (19).
\begin{exe}
    \item .....after civil-unrest in \textbf{Libya}...\\
    Potential event argument replacements from existing similar arguments from the corpus are \textbf{Iraq}, \textbf{Nigeria}, \textbf{Persian Gulf}, \textbf{Ukraine} and etc. \textbf{}
\end{exe}
\vspace{-1.0em}

\subsection{Analysis}
We have considered all annotation standards and their annotation methodologies and we have strived to align to those established by canonical programs. However, the annotation methodologies presented in this paper are adapted and changed to cater to special characteristics found in commodity news. For instance, under Event Metadata, \textit{Tense} and \textit{Genericity} found in ACE and ERE standards are dropped from our annotation scope while the new metadata - \textit{Intensity} is introduced. With accurate annotations, quality information can be extracted for other downstream tasks such as those listed in Section \ref{sec:Uses}. 

The commodity news dataset introduced here contains, in total, \textbf{3,950 events}, about \textbf{8,850 entities} and about \textbf{990 causal-precondition event-event relation}. Events distribution is shown is Table~\ref{table:EventDistribution}. Event Polarity, Modality and Intensity class distribution are presented in Table \ref{table:EventMPDistribution}.

\begin{table}[h!]   
    \centering
    \caption{Event type distribution and sentence level counts}
    \begin{tabular}{ l |r | c }  \hline
    \textbf{Event type} & \textbf{Type ratio} & \textbf{\# sentence instances} \\ \hline
    1. Cause-movement-down-loss & 13.35\% & 524  \\ \hline
    2. Cause-movement-up-gain & 2.23\% & 88 \\ \hline
    3. Civil-unrest & 2.53\% & 100\\ \hline
    4. Crisis & 0.76\% & 30 \\ \hline
    5. Embargo & 3.75\% & 148 \\ \hline
    6. Geopolitical-tension & 1.70\% & 67 \\ \hline
    7. Grow-strong & 6.03\% & 238 \\ \hline
    8. Movement-down-loss & 22.69\% & 896 \\ \hline
    9. Movement-flat & 1.52\% & 60 \\ \hline
    10. Movement-up-gain & 22.13\% & 874 \\ \hline
    11. Negative-sentiment & 4.79\% & 189 \\ \hline
    12. Oversupply & 2.63\% & 104 \\ \hline
    13. Position-high & 3.82\% & 151 \\ \hline
    14. Position-low & 3.11\% & 123 \\ \hline
    15. Prohibiting & 1.06\% & 42 \\ \hline
    16. Shortage & 1.04\% & 41 \\ \hline
    17. Slow-weak & 5.47\% & 216 \\ \hline
    18. Trade-tensions & 1.39\% & 55 \\ \hline \hline
    Total & & 3949 \\ \hline
    \end{tabular}
    \label{table:EventDistribution}
    \vspace{-0.7em}
\end{table}

 \begin{table}[h!]   
    \centering
    \caption{Event Polarity, Modality and Intensity Distribution}
    \begin{tabular}{ l | r | c }  \hline
    \textbf{Polarity} & \textbf{Type ratio} & \textbf{\# sentence instances} \\ \hline
    POSITIVE & 96.51\% & 3811  \\ \hline
    NEGATIVE & 3.49\% & 138 \\ \hline \hline
    \textbf{Modality} & \textbf{Type ratio} & \textbf{\# sentence instances} \\ \hline
    ASSERTED & 81.24\% & 3208 \\ \hline
    OTHER & 18.76\% & 741 \\ \hline \hline
    \textbf{Intensity} & \textbf{Type ratio} & \textbf{\# sentence instances} \\ \hline
    NEUTRAL & 87.59\% & 3459  \\ \hline
    EASED & 7.47\% & 295 \\ \hline
    INTENSIFIED & 4.94\% & 195 \\ \hline 
    \hline
    \end{tabular}
    \label{table:EventMPDistribution}
    \vspace{-0.7em}
\end{table}


\section{Evaluation and Annotation Challenges} \label{sec:challenges}
In total, the two annotators have annotated the same set of 150 documents and achieve a Cohen’s Kappa score of 0.787. This is very close to the near-perfect agreement range of [0.81, 0.99].


The main challenge of event annotation is ambiguities on `eventiveness'. Here are examples where information is conveyed in an indirect way that makes it difficult to pinpoint any clear-cut events:
        \begin{exe}
            \item Then spent the rest of the week trying to defend those gains as market optimism ......
            \item  Oil prices felt pressure on Tuesday from news that OPEC will slowly taper output cuts.
        \end{exe}
In cases above where an event trigger is absent, we do not annotate any event even though the event is conveyed in an implicit manner. 

The second challenge is the need for domain expertise to annotate the text. In many cases, it is imperative for annotators to understand financial and macro-economic terms and concepts to interpret the text accurately and annotate events accordingly. For instance, sentences containing macro-economic terms such as \textit{contango}, \textit{quantitative easing}, and \textit{backwardation} will require annotators to have finance and economics domain knowledge. To ensure that the annotators are knowledgeable with the domain knowledge, we recruited our annotators from a pool of undergraduate students from the School of Business.

The third challenge is the difference in interpreting special concepts that determines how these should be annotated. For example:
\begin{itemize}
    \item The word \textit{outlook}, should it be interpreted as \textit{forecast}? Or, should it be considered as a cue word for event modality?
    \item If events surrounding US \textit{employment} data are annotated, then what about \textit{unemployment}? Should this be treated as employment data but negated using negative polarity? 
    \item How should double negation be treated?  For example, `failed attempt to prevent a steep drop in oil prices', both \textit{failed} and \textit{prevent} are considered negative polarity cue words, creating a double negation situation.
\end{itemize}
For these non-straight forward cases, each one was handled on a case-by-case basis where the adjudicator discussed each situation with the annotators to seek consensus before finalizing an agreed annotation.


\section{Uses of the Corpus} \label{sec:Uses}
The main motivation for creating this annotated dataset is to contribute to resource building in the domain of finance and economics and to further facilitate research in this domain. Given the list of key information found in commodity news as described in Section \ref{intro}, the annotated information can potentially be used in the following tasks:
\begin{enumerate}
    \item \textbf{Event extraction}: One obvious use of this annotation is for event extraction. 
    \item \textbf{Event relation extraction}: Causal relation between events are useful in understanding the event chain, or known as script-learning in \citep{schank2013scripts}. This can be used to predict subsequent events given the occurrence of an event.
    \item \textbf{Nominal entity recognition in financial or macro-economics text}: A number of nominal entities relating to global macro-economic, geo-political and supply-demand situation were identified and duly annotated.  Full list of nominal entities is found in Appendix \ref{appendix:entityTypes}. This resource can be used for training a nominal entity recognition model as a complement component to existing Named Entity Recognition (NER) functionality. This will be help in processing, and analysing financial and economic text more effectively.
    \item \textbf{Polarity, modality and intensity classification}: Given that commodity corpus has abundant events with various combinations of polarity, modality and intensity (see sentence (12), (13) and (14) in Section \ref{subsection:metadata}), this dataset is a good resource for event polarity, modality and intensity classification task.
    \item \textbf{Potentially expandable to other types of commodities}: Event extraction model built using this dataset can potentially, through domain adaptation, be used for other commodities (such are gold, silver, palm oil and etc.) that are broadly influenced by the same types of global events as crude oil. 
\end{enumerate}

\section{Conclusion} \label{sec:Conclusion}
Event extraction in the domain of finance and economics at the moment are limited to only company-related events. To contribute to the building of resources in this domain, we have presented a new commodity news dataset with the following information annotated: (i) Entity mentions, (ii) Events (triggers and argument roles), (iii) Event Metadata (Polarity, Modality, and Intensity) and (iv) Event relations. We have also shared methodologies of how these information were annotated. Along with the dataset, we have also presented the various uses of the dataset. 

There are a number of avenues for future work. The main piece that can be further explored is to expand the annotation scope to cover more event types. Next, this work can also be expanded to cover more event-event relations such as event co-reference, sub-event, event-sequence, and contradictory event relation. Besides that, the current sentence-level annotation can be extended to cater for event relations spanning multiple sentences, so that event extraction and relation extract can be done at the document level.


\begin{acknowledgements}
Wharton Research Data Services (WRDS) was used in preparing this publication. This service and the data available thereon constitute valuable intellectual property and trade secrets of WRDS and/or its third-party suppliers.
\end{acknowledgements}

\appendix

\section{Ravenpack's Event Taxonomy}
\vspace{-1.0em}
RavenPack is an analytics provider for financial services. Among their services are finance and economic news sentiment analysis. Ravenpack has over 5000+ taxonomy categories and uses proprietary event matching methods to generate labels for their own news stream classifier. Based on news articles, they also calculate sentiment scores for commodity price prediction. Based on the analysis done by \citep{brandt2019macro}, Ravenpack's Event taxonomy is grouped into three broader classes: (i) Geo-political news, (ii) Macro-economic news, and (iii) Oil supply and demand news.

\label{appendix:eventTaxanomy}
\vspace{-2.0em}
\begin{table}[h!]
    \centering
        \caption{Categories of RavenPack event taxonomy grouped into three broader classes as presented in \citep{brandt2019macro}}.
        \begin{tabular}{ p{3.5cm} p{3.5cm} p{3.5cm}}
        \hline
        \multicolumn{3}{l}{\textbf{\underline{Geo-political News:}}} \\
        Terrorism &  War \& Conflict & Civil unrest \\ Natural disasters & Government &  \\ 
        \multicolumn{3}{l}{} \\
        \hline
        \multicolumn{3}{l}{\textbf{\underline{Macro-economic News:}}} \\
        Sovereign Debt & Public finance & Retail sales \\
        Consumer confidence & Housing & Interest rates \\ Treasury yield & Durable goods orders & Consumer spending \\
        Recession & Economic growth & GDP growth \\
        CPI & PPI & Trade balance \\
        Exports & Foreign exchange & Employment \\
        Private credit & & \\ 
        \multicolumn{3}{l}{} \\
        \hline
        \multicolumn{3}{l}{\textbf{\underline{Oil supply and demand:}}} \\
        Crude oil supply & Crude oil demand & Price Target \\
        Drilling \& pipeline accident & & \\
        \multicolumn{3}{l}{} \\
        \hline
        \end{tabular}
    \vspace{-2.0em}
    \label{table:RavenPackEventCategories}
\end{table}

\clearpage
\section{Entity Types} \label{appendix:entityTypes}
\vspace{-1.0em}
 \begin{table}[h!]   
    \begin{center}
    \caption{List of Entity Types}
    \begin{tabular}{ p{2.8cm} | p{8.2cm} }  \hline
    \textbf{Entity Type} & \textbf{Examples} \\ \hline
    1. Commodity & \textit{oil, crude oil, Brent, West Texas Intermediate (WTI), fuel, U.S Shale, light sweet crude, natural gas} \\ \hline
    2. Country** & \textit{Libya, China, U.S, Venezuela, Greece} \\ \hline
    3. Date** & \textit{1998, Wednesday, Jan. 30, the final quarter of 1991, the end of this year}  \\ \hline
    4. Duration** & \textit{two years, three-week, 5-1/2-year, multiyear, another six months} \\ \hline
    5. Economic Item & \textit{economy, economic growth, market, economic outlook, employment data, currency, commodity-oil}  \\ \hline
    6. Financial attribute & \textit{supply, demand, output, production, price, import, export} \\ \hline
    7. Forecast target & \textit{forecast, target, estimate, projection, bets} \\ \hline
    8. Group & \textit{global producers, oil producers, hedge funds, non-OECD, Gulf oil producers} \\ \hline
    9. Location** & \textit{global, world, domestic, Middle East, Europe} \\ \hline
    10. Money** & \textit{\$60, USD 50}  \\ \hline
    11. Nationality** & \textit{Chinese, Russian, European, African} \\ \hline
    12. Number** & (any numerical value that does not have a currency sign) \\ \hline
    13. Organization** & \textit{OPEC, Organization of Petroleum Exporting Countries, European Union, U.S. Energy Information Administration, EIA} \\ \hline
    14. Other activities & (free text) \\ \hline
    15. Percent** & \textit{25\%, 1.4 percent} \\ \hline
    16. Person** & \textit{Trump, Putin} (and other political figures) \\ \hline
    17. Phenomenon & (free text) \\ \hline
    18. Price unit & \textit{\$100-a-barrel, \$40 per barrel, USD58 per barrel}  \\ \hline
    19. Production Unit & \textit{170,000 bpd, 400,000 barrels per day, 29 million barrels per day} \\ \hline
    20. Quantity & \textit{1.3500 million barrels, 1.8 million gallons, 18 million tonnes} \\ \hline
    21. State or province** & \textit{Washington, Moscow, Cushing, North America} \\ \hline
    \end{tabular}
    \vspace{-1.0em}
    \label{table:EntityTypes}
    \end{center}
\end{table}

Entity type marked with ** in Table~\ref{table:EntityTypes} are made up of mostly Named Entities. These named entities are identitical to the NER-tagging in Stanford CoreNLP.

\clearpage

\section{Event Types} \label{appendix:eventTypes}
The list of annotated events are aligned to Ravenpack's event taxanomy categeories which is made publicly available. There are 18 event types identified and annotated, the event types are listed in Table~\ref{table:EventTypes} below:
\vspace{-0.7em}
 \begin{table}[h!]   
    \centering
    \caption{List of Event Types}
    \begin{tabular}{ p{3.5cm} | p{7.5cm} }  \hline
    \textbf{Event Type} & \textbf{Example Trigger Word(s)} \\ \hline
    1. Cause-movement-down-loss & \textit{cut, trim, reduce, disrupt, curb, squeeze, choked off} \\ \hline
    2. Cause-movement-up-gain & \textit{boost, revive, ramp up, prop up, raise} \\ \hline
    3. Civil-unrest & \textit{violence, turmoil, fighting, civil war, insurgent attacks, conflicts}  \\ \hline
    4. Crisis & \textit{crisis, crises} \\ \hline
    5. Embargo & \textit{embargo, sanction}  \\ \hline
    6. Geopolitical-tension & \textit{war, clashes, tensions, deteriorating relationship} \\ \hline
    7. Grow-strong & \textit{grow, picking up, boom, recover, expand, strong, rosy, improve, solid} \\ \hline
    8. Movement-down-loss & \textit{fell, down, less, drop, tumble, collapse, plunge, downturn, slump, slide, decline} \\ \hline
    9. Movement-flat & \textit{unchanged, flat, hold, no change, maintained} \\ \hline
    10. Movement-up-gain & \textit{up, gain, rise, surge, soar, swell, increase, rebound} \\ \hline
    11. Negative-sentiment & \textit{worries, concern} \\ \hline
    12. Oversupply & \textit{ample supply, glut, oversupply, bulging stock level, excess supplies} \\ \hline
    13. Position-high & \textit{high, highest, peak, highs} \\ \hline
    14. Position-low & \textit{low, lowest, lows, trough} \\ \hline
    15. Prohibiting & \textit{ban, bar, prohibit} \\ \hline
    16. Shortage & \textit{shortfall, shortage, under-supplied} \\ \hline
    17. Slow-weak & \textit{slow, weak, tight, lackluster, falter, weaken, bearish, slowdown, crumbles} \\ \hline
    18. Trade-tensions & \textit{price war, trade war, trade tensions, economic fallout, trade dispute} \\ \hline
    \end{tabular}
    \vspace{-1.0em}
    \label{table:EventTypes}
\end{table}

As described in Section \ref{intro}, events are not fully represented if we just carry out Event Detection focusing on only event triggers. The events listed in Table \ref{table:EntityTypes} can be grouped into the following categories by coupling event arguments with event triggers:
\subsection{Geo-political News}
\begin{enumerate}
    \item Civil Unrest (\textbf{Civil-unrest}) - Violence or turmoil within the oil producing country
        \begin{exe}
        \item .....a fragile recovery in Libyan supply outweighed \textbf{fighting} in Iraq ......
        \item .......a backdrop of the worst \textbf{strife} in Iran this decade....
        \end{exe}
    \item Embargo (\textbf{Embargo, Prohibiting}) - Trade or other commercial activity of the commodity is banned.
        \begin{exe}
            \item ..... and \textbf{sanctions} against Iran.
            \item .....prepared to impose `` strong and swift '' economic \textbf{sanctions} on Venezuela.....
        \end{exe}
    \item Geo-political Tension (\textbf{Geo-political-tension}) - Political tension between oil-producing nation with other nations.
        \begin{exe}
            \item ..... heightened \textbf{tensions} between the West and Russia.....
            \item ..... despite geopolitical \textbf{war} in Iraq , Libya and Ukraine.
        \end{exe}
    \item Trade Tension (\textbf{Trade-tensions}) - Trade-related tension between oil-producing and oil-consuming nations.
        \begin{exe}
            \item  ..... escalating global \textbf{trade wars}, especially between the US and China.
            \item .... showing that OPEC is not ready to end its \textbf{trade tensions}......
        \end{exe}
    \item Other forms of Crisis (\textbf{Crisis}) 
        \begin{enumerate}
            \item A time of intense difficulty, such as other forms of unspecified crisis that do not fall into any of the above category
                \begin{exe}
                    \item  .... Ukraine declared an end to an oil \textbf{crisis} that has .........
                \end{exe}
            \item Financial / Economic Crisis (which can be grouped under Macro-economic News) 
                \begin{exe}
                    \item ....since the 2014/15 financial \textbf{crisis} as .......
                \end{exe}
        \end{enumerate}
\end{enumerate}

\subsection{Macro-economic News}
\begin{enumerate}
    \item Strong Economy / GDP growth / US Employment (\textbf{Grow-strong}) - Strong or growing economy / GDP of a nation; applicable to indicate strong status of US Employment Data \footnote{US Employment data is used as an indicator of the world largest's economic health}.        
        \begin{exe}
            \item as \textbf{strong} U.S. employment data.....
        \end{exe}
    \item Weak or Contracting Economy / GDP / US Employment (\textbf{Slow-weak}) - Weakening or contracting economy / GDP of a nation; applicable to indicate the weakening of US employment data. 
        \begin{exe}
            \item  ....... concerns over \textbf{slowing} global growth.....
            \item U.S. employment data \textbf{contracts} with the euro zone.....
        \end{exe}
    \item Bearish technical view or outlook (\textbf{Negative-sentiment}) - Bearish sentiment or outlook
        \begin{exe}
            \item But in a market \textbf{clouded by uncertainties}.....
            \item ....supply \textbf{concerns} would ease even more ......
        \end{exe}
\end{enumerate}

\subsection{Commodity Supplies including exports}
    \begin{enumerate}
        \item Oversupply (\textbf{Oversupply}) - Situation where production goes into surplus.
            \begin{exe}
                \item ..... the region \textbf{surplus} of supply.....
                \item ....the market is still working off the \textbf{gluts} built up.....
            \end{exe}
        \item Shortage (\textbf{Shortage}) - Situation where demand is more than supply.
            \begin{exe}
                \item .... increase a supply \textbf{shortage} from chaotic Libya....
                \item ......and there is no \textbf{shortfall} in supply , the minister added.
            \end{exe}
        \item Supply increase (\textbf{Movement-up-gain}) - Situation where supply increased.
            \begin{exe}
                \item ....further \textbf{increases} in U.S. crude production.....
                \item The \textbf{rise} in production is definitely benefiting the United States....
            \end{exe}
        \item Action taken to increase supply (\textbf{Cause-Movement-Up-gain}) - Deliberate action to increase supply.
        \begin{exe}
            \item The IEA \textbf{boosted} its estimate of production from ExxonMobil to 1.8 million bpd in July 4 holiday weekend.
            \item .....urged the kingdom to \textbf{ramp up} production.....
        \end{exe}
        \item Supply decrease (\textbf{Movement-down-loss})- Situation where supply decreased.
        \begin{exe}
            \ex UAE 's production has almost \textbf{halved} in two years to 31.6 million bpd...
            \ex ...fears that global supplies will \textbf{drop} due to Washington 's sanctions on the OPEC member nation.
        \end{exe}
        \item Action taken to decrease supply (\textbf{ause-movement-down-loss}) - Deliberate action to decrease supply.
        \begin{exe}
            \ex ......by \textbf{slashing} production by almost three quarters in the 1980s....
            \ex .....an announcement by Iran that it would \textbf{cut} its production last week.
        \end{exe}
    \end{enumerate}
    
\subsection{Commodity Demand including imports}
    \begin{enumerate}
        \item Demand increase (\textbf{Movement-up-gain}) - Situation where demand increased.
        \begin{exe}
            \ex It expects consumption to \textbf{trend upward} by 1.05 million bpd , below 40,000 bpd from July .
            \ex ....as \textbf{more} seasonal demand kicks in due to colder weather.
        \end{exe}
        \item Action taken to increase demand (\textbf{Caused-movement-up-gain}) - Deliberate action taken to increase demand.
        \begin{exe}
            \ex IEA tried to \textbf{boost} global oil demand by introducing......
        \end{exe}
        \item Demand decrease (\textbf{Movement-down-loss}) - Situation where demand decreased.
        \begin{exe}
            \ex ....onto a market reeling from \textbf{falling} demand because of the virus outbreak.
            \ex ....when global demand growth for air conditioning \textbf{collapses} from its summer peak....
        \end{exe}
        \item Action taken to decrease demand (\textbf{Caused-movement-down-loss})- Deliberate action taken to decrease demand.
        \begin{exe}
            \ex The pandemic has \textbf{zapped} demand to a level never seen before...
        \end{exe}
    \end{enumerate}

\subsection{Commodity Price Movement}
Commodity price here includes \textit{spot price}, \textit{futures} and \textit{futures contract}
    \begin{enumerate}
        \item Price increase (\textbf{Movement-up-gain})- Situation where commodity price rises.
        \begin{exe}
            \ex Oil price \textbf{rose} \$105 a barrel on March....
            \ex ...oil prices have \textbf{jumped} as much as 20 percent since June.
        \end{exe}
        \item Price decrease (Movement-down-loss) - Situation where commodity price drops.
        \begin{exe}
            \ex The \textbf{drop} in oil prices to their lowest in two years.....
            \ex Oil prices \textbf{declined} back the final quarter of 1991 to 87 cents....
        \end{exe}
        \item Price movement flat (\textbf{Movement-flat}) - Situation where no or little change to commodity price.
        \begin{exe}
            \ex Contango spread in Brent is steady at 15 cents per barrel.....
            \ex U.S. crude is expected to \textbf{hold} around \$105 per barrel.
        \end{exe}
        \item Price position (\textbf{Position-high, Position-low}) - Describes the position of the current commodity price.
        \begin{exe}
            \ex Oil price remained close to four-year \textbf{highs}....
            \ex Oil slipped more than 20\% to its \textbf{lowest level} in two years on 1980s....
        \end{exe}
    \end{enumerate}

\subsection{Change in Forecasted value}
    \begin{enumerate}
        \item Increase forecast target (\textbf{Caused-movement-up-gain}) - Forecasted / target is raised, possible values are price target, growth target, demand and supply target.
        \begin{exe}
            \ex The IMF earlier said it \textbf{increased} its 2019 global economic growth forecast to 3.30\%.
            \ex The International Monetary Fund \textbf{doubled} its global growth forecast for 2013.....
        \end{exe}
        \item Price target /forecast decrease (\textbf{Caused-movement-down-loss}) - Forecasted / target is lowered, possible values are commodity price target, growth target, demand and supply target.
        \begin{exe}
            \ex Germany's Bundesbank this week \textbf{halved} its 2015 growth forecasts for Europe 's largest economy to 1 percent.
            \ex OPEC also \textbf{lowered} forecast global demand for its crude oil....
        \end{exe}
    \end{enumerate}
    
\clearpage
\section{Event Arguments} \label{appendixEventArguments}
\subsection{Movement-down-loss, Movement-up-gain, Movement-flat}
\textbf{Example sentence: } [Globally] [crude oil] [futures] \textbf{surged} [\$2.50] to [\$59 per barrel] on [Tuesday].
\vspace{-2.0em}
 \begin{table}[h]   
    \centering
    \begin{tabular}{ | p{2.9cm} | p{6.5cm} | p{2.3cm} |}  \hline
    \textbf{Role} & \textbf{Entity Type} & \textbf{Argument Text}\\ \hline
    
     Type & Nationality, Location & globally \\ \hline
     Place & Country, Group, Organization, Location, State or province, Nationality & \\ \hline
     Supplier\_consumer & Organization, Country, State\_or\_province, Group, Location & \\ \hline
     Reference\_point\_time & Date & Tuesday \\ \hline
     Initial\_reference\_point & Date &  \\ \hline
     Final\_value & Percentage, Number, Money, Price\_unit, Production\_unit, Quantity & \$59 per barrel \\ \hline
     Initial\_value & Percentage, Number, Money, Price\_unit, Production\_unit, Quantity &  \\ \hline
     Item & Commodity, Economic\_item & crude oil \\ \hline
     Attribute & Financial\_attribute & futures\\ \hline
     Difference & Percentage, Number, Money, Production\_unit, Quantity & \$2.50 \\ \hline
     Forecast & Forecast\_target &  \\ \hline
     Duration & Duration & \\ \hline
     Forecaster & Organization & \\ \hline 
    \end{tabular}

    \vspace{-2.5em}
\end{table}

\subsection{Embargo}
\textbf{Example sentence:} The [Trump administration] imposed a ``strong and swift'' economic \textbf{sanctions} on [Venezuela] on [Thursday].
\vspace{-1.0em}
 \begin{table}[h]   
    \centering
    \begin{tabular}{ | p{2.9cm} | p{6.5cm} | p{2.3cm} |}  \hline
    \textbf{Role} & \textbf{Entity Type} & \textbf{Argument Text}\\ \hline
     Imposer & Organization, Country, Nationality, State or province, Person, Group, Location &  Trump administration \\ \hline
     Imposee\footnote{Not formally a word, but used here as a shorter version of ``Party whom the action was imposed on} & Organization, Country, Nationality, State or province, Group & Venezuela \\ \hline
     Reference\_point\_time & Date & Thursday \\ \hline
    \end{tabular}
    \vspace{-2.5em}
\end{table}

\subsection{Prohibiting}
\textbf{Example sentence:} [Congress] \textbf{banned} most [U.S.] [crude oil] [exports] on [Friday] after price shocks from the 1973 Arab oil embargo.
\vspace{-1.0em}
 \begin{table}[h]   
    \centering
    \begin{tabular}{ | p{2.9cm} | p{6.5cm} | p{2.3cm} |}  \hline
    \textbf{Role} & \textbf{Entity Type} & \textbf{Argument Text}\\ \hline
     Imposer & Organization, Country, Nationality, State or province, Person, Group, Location &  Congress \\ \hline
     Imposee & Organization, Country, Nationality, State or province, Group & U.S. \\ \hline
     Item & Commodity, Economic\_item & crude oil \\ \hline
     Attribute & Financial\_attribute & exports \\ \hline
     Reference\_point\_time & Date & Friday \\ \hline
     Activity & Other\_activities & \\ \hline
    \end{tabular}
    \vspace{-2.5em}
\end{table}

\subsection{Caused-movement-down-loss, Caused-movement-up-gain}
\textbf{Example sentence:} The [IMF] earlier said it \textbf{reduced} its [2018] [global] [economic growth] [forecast] to [3.30\%] from a [July] forecast of [4.10\%].
\vspace{-1.0em}
 \begin{table}[h]
    \centering
    \begin{tabular}{ | p{2.9cm} | p{6.5cm} | p{2.3cm} |}  \hline
    \textbf{Role} & \textbf{Entity Type} & \textbf{Argument Text}\\ \hline
    
     Type & Nationality, Location & global \\ \hline
     Place & Country, Group, Organization, Location, State or province, Nationality & West African, European \\ \hline
     Supplier\_consumer & Organization, Country, State\_or\_province, Group, Location & \\ \hline
     Reference\_point\_time & Date & 2018 \\ \hline
     Initial\_reference\_point & Date & July \\ \hline
     Final\_value & Percentage, Number, Money, Price\_unit, Production\_unit, Quantity & 3.30\%\\ \hline
     Initial\_value & Percentage, Number, Money, Price\_unit, Production\_unit, Quantity & 4.10\% \\ \hline
     Item & Commodity, Economic\_item & economic growth \\ \hline
     Attribute & Financial\_attribute & \\ \hline
     Difference & Percentage, Number, Money, Production\_unit, Quantity & \\ \hline
     Forecast & Forecast\_target & forecast \\ \hline
     Duration & Duration & \\ \hline
     Forecaster & Organization & IMF \\ \hline 
    \end{tabular}
    \vspace{-2.0em}
\end{table}
    
\subsection{Shortage}
\textbf{Example Sentence:} Oil reserves are within ``acceptable'' range in most oil consuming countries and there is no \textbf{shortage} in [oil] [supply] [globally], the minister added.
\vspace{-1.0em}
 \begin{table}[h]   
    \centering
    \begin{tabular}{ | p{2.9cm} | p{6.5cm} | p{2.3cm} |}  \hline
    \textbf{Role} & \textbf{Entity Type} & \textbf{Argument Text}\\ \hline
     Place & Country, State or province, Location, Nationality &  Congress \\ \hline
     Item & Commodity & crude oil \\ \hline
     Attribute & Financial\_attribute & exports \\ \hline
     Type & Location & globally \\ \hline
     Reference\_point\_time & Date &  \\ \hline
    \end{tabular}
    \vspace{-2.0em}
\end{table}

\subsection{Oversupply}
\textbf{Example sentence:} [Forecasts] for an [crude] \textbf{oversupply} in [West African] and [European] [markets] [early June] help to push the Brent benchmark down more than 20\% January.
\vspace{-1.0em}
 \begin{table}[h]   
    \centering
    \begin{tabular}{ | p{2.9cm} | p{6.5cm} | p{2.3cm} |}  \hline
    \textbf{Role} & \textbf{Entity Type} & \textbf{Argument Text}\\ \hline
     Place & Country, Group, Organization, Location, State or province, Nationality & West African, European \\ \hline
     Reference\_point\_time & Date & this year \\ \hline
     Item & Commodity & crude \\ \hline
     Attribute & Financial\_attribute & markets \\ \hline
     Difference & Production\_unit & \\ \hline
     Forecast & Forecast\_target & forecasts \\ \hline
    \end{tabular}
    \vspace{-2.0em}
\end{table}

\subsection{Position-high, Position-low}
\textbf{Example sentence:} The IEA estimates that U.S. crude oil is expected to seek higher ground until reaching a [5-year] \textbf{peak} in [late April] of about [17 million bpd].
\vspace{-1.0em}
 \begin{table}[h!]   
    \centering
    \begin{tabular}{ | p{2.9cm} | p{6.5cm} | p{2.3cm} |}  \hline
    \textbf{Role} & \textbf{Entity Type} & \textbf{Argument Text}\\ \hline
         Reference\_point\_time & Date &  late April \\ \hline
     Initial\_reference\_point & Date &  \\ \hline
     Final\_value & Percentage, Number, Money, Price\_unit, Production\_unit, Quantity &  17 million bpd\\ \hline
     Initial\_value & Percentage, Number, Money, Price\_unit, Production\_unit, Quantity &  \\ \hline
     Item & Commodity, Economic\_item & \\ \hline
     Attribute & Financial\_attribute & \\ \hline
     Difference & Percentage, Number, Money, Production\_unit, Quantity &  \\ \hline
     Duration & Duration & 5-year \\ \hline
    \end{tabular}
    \vspace{-2.0em}
\end{table}

\subsection{Geo-political Tension}
\textbf{Example sentence: } \textbf{Deteriorating relations} between [Iraq] and [Russia] [first half of 2016] ignited new fears of supply restrictions in the market.
\vspace{-1.0em}
\textbf{Trade Tension}
 \begin{table}[h]   
    \centering
    \begin{tabular}{ | p{2.9cm} | p{6.5cm} | p{2.3cm} |}  \hline
    \textbf{Role} & \textbf{Entity Type} & \textbf{Argument Text}\\ \hline
     Participating\_countries & Country, Group, Organization, Location, State or province, Nationality & U.S., China \\ \hline
     Reference\_point\_time & Date & early June \\ \hline
    \end{tabular}
    \vspace{-2.5em}
\end{table}

\subsection{Slow-weak, Grow-strong}
\textbf{Example sentence:}  [U.S.] [employment data] \textbf{strengthens} with the euro zone. 
\vspace{-1.0em}
 \begin{table}[h]   
    \centering
    \begin{tabular}{ | p{2.9cm} | p{6.5cm} | p{2.3cm} |}  \hline
    \textbf{Role} & \textbf{Entity Type} & \textbf{Argument Text}\\ \hline
    
     Type & Nationality, Location & \\ \hline
     Place & Country, Group, Organization, Location, State or province, Nationality & U.S. \\ \hline
     Supplier\_consumer & Organization, Country, State\_or\_province, Group, Location & \\ \hline
     Reference\_point\_time & Date &  \\ \hline
     Initial\_reference\_point & Date &  \\ \hline
     Final\_value & Percentage, Number, Money, Price\_unit, Production\_unit, Quantity &  \\ \hline
     Initial\_value & Percentage, Number, Money, Price\_unit, Production\_unit, Quantity &  \\ \hline
     Item & Commodity, Economic\_item & employment data \\ \hline
     Attribute & Financial\_attribute & \\ \hline
     Difference & Percentage, Number, Money, Production\_unit, Quantity &  \\ \hline
     Forecast & Forecast\_target &  \\ \hline
     Duration & Duration & \\ \hline
     Forecaster & Organization & \\ \hline 
    \end{tabular}
    \vspace{-1.5em}    
\end{table}

\subsection{Civil Unrest}
\textbf{Example sentence:} The drop in oil prices to their lowest in two years has caught many observers off guard, coming against a backdrop of the worst \textbf{violence} in [Iraq] [this decade].
    \vspace{-1.0em}
 \begin{table}[h]   
    \centering
    \begin{tabular}{ | p{2.9cm} | p{6.5cm} | p{2.3cm} |}  \hline
    \textbf{Role} & \textbf{Entity Type} & \textbf{Argument Text}\\ \hline
     Place & Country, State or province, Location, Nationality & Iraq  \\ \hline
     Reference\_point\_time & Date & this decade \\ \hline
    \end{tabular}
    \vspace{-2.0em}
\end{table}

\subsection{Crisis}
\textbf{Example Sentence:} Asia 's diesel consumption is expected to recover this year at the second weakest level rate since the [2014] [Asian] [financial] \textbf{crisis}.
\vspace{-1.0em}
 \begin{table}[h]   
    \centering
    \begin{tabular}{ | p{2.9cm} | p{6.5cm} | p{2.3cm} |}  \hline
    \textbf{Role} & \textbf{Entity Type} & \textbf{Argument Text}\\ \hline
     Place & Country, State or province, Location, Nationality & Asian \\ \hline
     Reference\_point\_time & Date & this year \\ \hline
     Item & Commodity, Economic\_item & financial \\ \hline 
    \end{tabular}
    \vspace{-2.0em}
\end{table}

\subsection{Negative Sentiment}
\textbf{Example sentence:} Oil futures have dropped due to \textbf{concern} about softening demand growth and awash in crude.

Note: \textbf{Negative Sentiment} is a special type of event, where majority of the time it contains just the trigger words such as \textit{concerns, worries, fears} and 0 event arguments.


\clearpage
\bibliographystyle{spbasic}      
\bibliography{bibtex}

\begin{thebibliography}{14}
\providecommand{\natexlab}[1]{#1}
\providecommand{\url}[1]{{#1}}
\providecommand{\urlprefix}{URL }
\expandafter\ifx\csname urlstyle\endcsname\relax
  \providecommand{\doi}[1]{DOI~\discretionary{}{}{}#1}\else
  \providecommand{\doi}{DOI~\discretionary{}{}{}\begingroup
  \urlstyle{rm}\Url}\fi
\providecommand{\eprint}[2][]{\url{#2}}

\bibitem[{Aguilar et~al(2014)Aguilar, Beller, McNamee, Van~Durme, Strassel,
  Song, and Ellis}]{aguilar2014comparison}
Aguilar J, Beller C, McNamee P, Van~Durme B, Strassel S, Song Z, Ellis J (2014)
  A comparison of the events and relations across ace, ere, tac-kbp, and
  framenet annotation standards. In: Proceedings of the Second Workshop on
  EVENTS: Definition, Detection, Coreference, and Representation, pp 45--53

\bibitem[{Araki et~al(2018)Araki, Mulaffer, Pandian, Yamakawa, Oflazer, and
  Mitamura}]{araki-etal-2018-interoperable}
Araki J, Mulaffer L, Pandian A, Yamakawa Y, Oflazer K, Mitamura T (2018)
  Interoperable annotation of events and event relations across domains. In:
  Proceedings 14th Joint {ACL} - {ISO} Workshop on Interoperable Semantic
  Annotation, Association for Computational Linguistics, Santa Fe, New Mexico,
  USA, pp 10--20

\bibitem[{Brandt and Gao(2019)}]{brandt2019macro}
Brandt MW, Gao L (2019) Macro fundamentals or geopolitical events? a textual
  analysis of news events for crude oil. Journal of Empirical Finance 51:64--94

\bibitem[{Dor et~al(2019)Dor, Gera, Toledo-Ronen, Halfon, Sznajder, Dankin,
  Bilu, Katz, and Slonim}]{dor2019financial}
Dor LE, Gera A, Toledo-Ronen O, Halfon A, Sznajder B, Dankin L, Bilu Y, Katz Y,
  Slonim N (2019) Financial event extraction using wikipedia-based weak
  supervision. In: Proceedings of the Second Workshop on Economics and Natural
  Language Processing, pp 10--15

\bibitem[{Hogenboom et~al(2013)Hogenboom, Hogenboom, Frasincar, Schouten, and
  Van Der~Meer}]{hogenboom2013semantics}
Hogenboom A, Hogenboom F, Frasincar F, Schouten K, Van Der~Meer O (2013)
  Semantics-based information extraction for detecting economic events.
  Multimedia Tools and Applications 64(1):27--52

\bibitem[{Jacobs et~al(2018)Jacobs, Lefever, and Hoste}]{jacobs2018economic}
Jacobs G, Lefever E, Hoste V (2018) Economic event detection in
  company-specific news text. In: The 56th Annual Meeting of the Association
  for Computational Linguistics, Association for Computational Linguistics, pp
  1--10

\bibitem[{Malik et~al(2011)Malik, Bhardwaj, and Fiorletta}]{malik2011accurate}
Malik HH, Bhardwaj VS, Fiorletta H (2011) Accurate information extraction for
  quantitative financial events. In: Proceedings of the 20th ACM international
  conference on information and knowledge management, pp 2497--2500

\bibitem[{Mitamura et~al(2015)Mitamura, Yamakawa, Holm, Song, Bies, Kulick, and
  Strassel}]{mitamura2015event}
Mitamura T, Yamakawa Y, Holm S, Song Z, Bies A, Kulick S, Strassel S (2015)
  Event nugget annotation: Processes and issues. In: Proceedings of the The 3rd
  Workshop on EVENTS: Definition, Detection, Coreference, and Representation,
  pp 66--76

\bibitem[{O{'}Gorman et~al(2016)O{'}Gorman, Wright-Bettner, and
  Palmer}]{ogorman-etal-2016-richer}
O{'}Gorman T, Wright-Bettner K, Palmer M (2016) Richer event description:
  Integrating event coreference with temporal, causal and bridging annotation.
  In: Proceedings of the 2nd Workshop on Computing News Storylines ({CNS}
  2016), Association for Computational Linguistics, Austin, Texas, pp 47--56

\bibitem[{O’Gorman et~al(2016)O’Gorman, Wright-Bettner, and
  Palmer}]{o2016richer}
O’Gorman T, Wright-Bettner K, Palmer M (2016) Richer event description:
  Integrating event coreference with temporal, causal and bridging annotation.
  In: Proceedings of the 2nd Workshop on Computing News Storylines (CNS 2016),
  pp 47--56

\bibitem[{Schank and Abelson(2013)}]{schank2013scripts}
Schank RC, Abelson RP (2013) Scripts, plans, goals, and understanding: An
  inquiry into human knowledge structures. Psychology Press

\bibitem[{Stenetorp et~al(2012)Stenetorp, Pyysalo, Topi{\'c}, Ohta, Ananiadou,
  and Tsujii}]{stenetorp2012brat}
Stenetorp P, Pyysalo S, Topi{\'c} G, Ohta T, Ananiadou S, Tsujii J (2012) Brat:
  a web-based tool for nlp-assisted text annotation. In: Proceedings of the
  Demonstrations at the 13th Conference of the European Chapter of the
  Association for Computational Linguistics, pp 102--107

\bibitem[{Thompson et~al(2017)Thompson, Nawaz, McNaught, and
  Ananiadou}]{thompson2017enriching}
Thompson P, Nawaz R, McNaught J, Ananiadou S (2017) Enriching news events with
  meta-knowledge information. Language Resources and Evaluation 51(2):409--438

\bibitem[{Yang et~al(2018)Yang, Chen, Liu, Xiao, and
  Zhao}]{yang-etal-2018-dcfee}
Yang H, Chen Y, Liu K, Xiao Y, Zhao J (2018) {DCFEE}: A document-level
  {C}hinese financial event extraction system based on automatically labeled
  training data. In: Proceedings of {ACL} 2018, System Demonstrations,
  Association for Computational Linguistics, Melbourne, Australia, pp 50--55

\end{thebibliography}

\end{document}